\documentclass[journal]{IEEEtran}
\usepackage{lipsum}
\usepackage[mathlines,switch]{lineno}
\usepackage{amsmath}

\DeclareMathOperator{\tr}{tr}
\DeclareMathOperator{\diag}{diag}

\usepackage{cite}

\ifCLASSINFOpdf
  \usepackage[pdftex]{graphicx}
\else
   \usepackage[dvips]{graphicx}
\fi
\usepackage{amsmath}
\usepackage{bm}
\usepackage{algorithm,algorithmic}

\usepackage{booktabs}
\usepackage{array}
\ifCLASSINFOpdf
\else
\fi
\usepackage{framed,multirow}
\usepackage{color}
\usepackage{lineno}
\usepackage{dblfloatfix}

\begin{document}
\title{Pairwise Point Cloud Registration Using Graph Matching and Rotation-invariant Features}

%
\author{Rong~Huang,~\IEEEmembership{Student~Member,~IEEE,}
        Wei~Yao,
        Yusheng~Xu,~\IEEEmembership{Member,~IEEE,}
        Zhen~Ye,
        and~Uwe~Stilla,~\IEEEmembership{Senior~Member,~IEEE}
\thanks{\emph{Corresponding author: Wei Yao.}}
\thanks{R. Huang, Y. Xu, and U. Stilla are with Photogrammetry and Remote Sensing, Technical University of Munich, 80333 Munich, Germany. (e-mail: rong.huang@tum.de;yusheng.xu@tum.de;stilla@tum.de).}
\thanks{W. Yao is with Department of Land Surveying and Geo-Informatics, The Hong Kong Polytechnic University, Hung Hom, Hong Kong. (e-mail: wei.hn.yao@polyu.edu.hk).}
\thanks{Z. Ye is with College of Surveying and Geo-Informatics, Tongji University, Shanghai, China. (e-mail: 89{\_}yezhen@tongji.edu.cn).}
\thanks{This work has been submitted to the IEEE for possible publication. Copyright may be transferred without notice, after which this version may no longer be accessible. Manuscript received MM DD, YYYY.}}

%


\maketitle

\begin{abstract}
Registration is a fundamental but critical task in point cloud processing, which usually depends on finding element correspondence from two point clouds.
However, the finding of reliable correspondence relies on establishing a robust and discriminative description of elements and the correct matching of corresponding elements. 
In this letter, we develop a coarse-to-fine registration strategy, which utilizes rotation-invariant features and a new weighted graph matching method for iteratively finding correspondence. In the graph matching method, the similarity of nodes and edges in Euclidean and feature space are formulated to construct the optimization function. The proposed strategy is evaluated using two benchmark datasets and compared with several state-of-the-art methods. Regarding the experimental results, our proposed method can achieve a fine registration with rotation errors of less than 0.2 degrees and translation errors of less than 0.1 m.
\end{abstract}


%
\IEEEpeerreviewmaketitle

\section{Introduction}

\IEEEPARstart{R}{e}gistration of point clouds plays an essential role in the community of photogrammetry and 3D computer vision, serving as a prerequisite for many applications such as change detection, dynamic monitoring, data fusion, and pose estimation \cite{dong2020registration}. 
Concisely,  the registration is to unite 3D point clouds acquired from various views, platforms, or times into a common coordinate system. 
Here, the core of achieving this objective is to correspond identical elements in a local or global way.
Generally, the commonly used strategy for registration methods mainly consists of at least two essential steps: extracting and describing feature elements and finding the correspondence between elements, with which transformation parameters are estimated. 
Regarding the extraction of element features and finding correspondence, there are a couple of underlying requirements to meet: 
(i) The extracted features should be invariant to transformations like rotation and translation; 
(ii) The feature description should be insensitive to uneven densities, noise, and outliers;
(iii) The finding of correspondence should be efficient and available for large data;
(iv) The transformation using found correspondence should be robust to incorrect matches.

Considering these requirements, many solutions have been proposed coping with the element feature extraction and description and the finding of correspondence. 
Features are essential in correspondence matching, through which the similarity measurement between features indicates identical points. 
Popular feature descriptors include scale-invariant feature transform (SIFT) \cite{flitton2010object}, fast point feature histogram (FPFH) \cite{rusu2009fast}, and signature of histogram of orientations (SHOT) \cite{tombari2010unique} and so on. 
Descriptiveness and rotation-invariance are important criteria for a competent feature descriptor. 
The descriptiveness highly depends on the descriptor's performance and the geometry of described area, which changes from methods to methods. However, the rotation-invariance is a must for all the descriptors.  
A general strategy for achieving rotation in-variance is based on pose normalization using a local reference frame (LRF) \cite{tombari2010unique}. 
However, most of the LRF in descriptors suffer from noise, outliers, and occluded surfaces.
This presents a requirement of robustness for feature descriptors.
Apart from features of points, many other researches utilized geometric primitives as features instead of points, such as lines \cite{habib2005photogrammetric}, curves \cite{yang2014automated}, planes \cite{chen2019plade}. 
Such primitives have fewer degrees of freedom in orientation, providing additional constraints \cite{xu2019pairwise}.
\begin{figure*}[t!]
    \centering
    \includegraphics[width=0.9\textwidth]{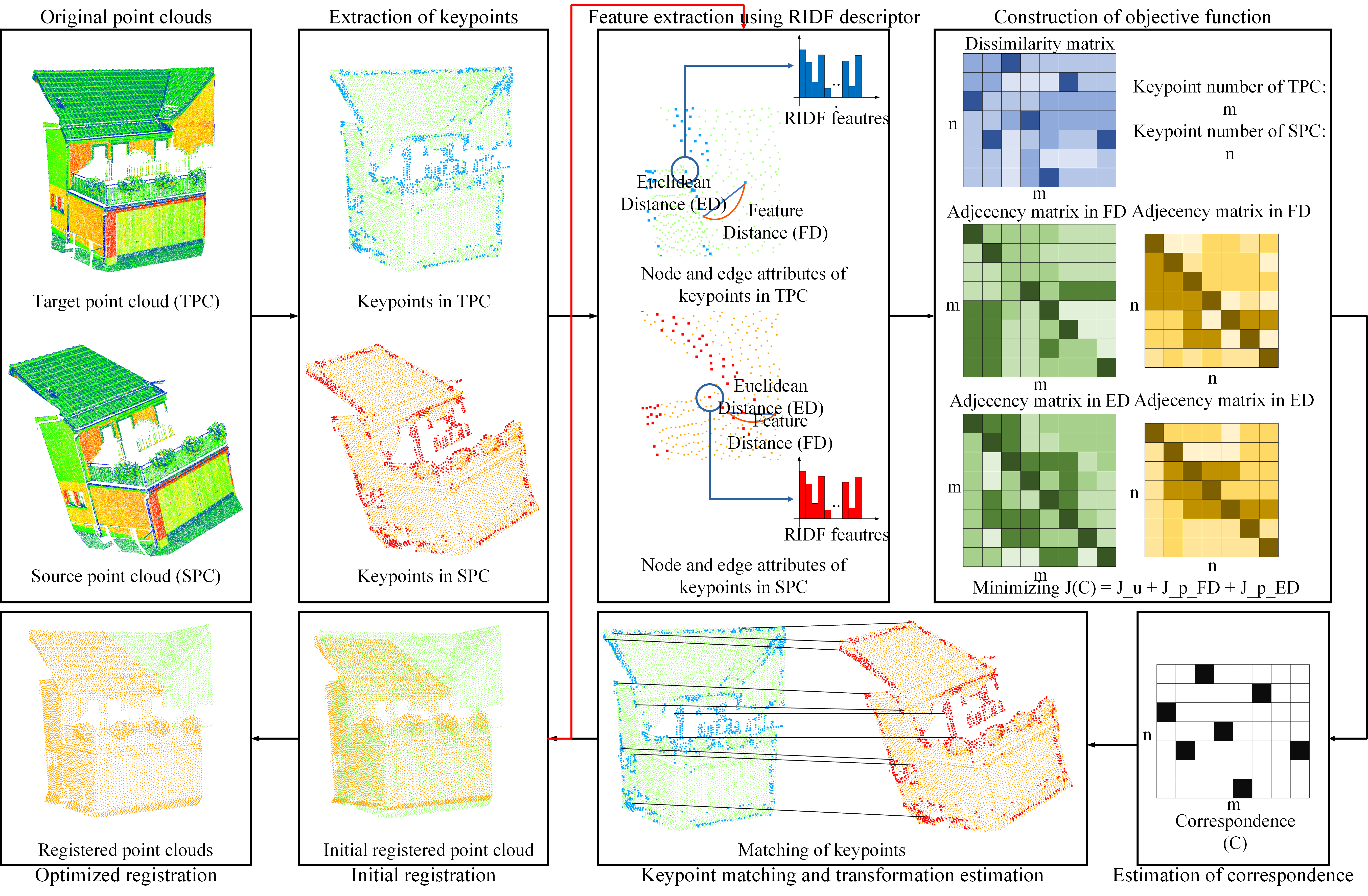}
    \caption{Overview of workflow of the proposed registration method.}
    \label{fig:FigX_workflow}
\end{figure*}

To identify correspondence, random sample consensus (RANSAC) and 4-points congruent set (4PCS) \cite{aiger20084} are renowned methods to find the maximum congruence. However, both of them may meet efficiency problems \cite{ge2017automatic}. 
Guided sampling \cite{quan2020compatibility} is an improvement utilizing constraints from features to guide the sampling process, which can accelerate the finding of congruence.  
Apart from sampling consensus-based methods, bi-graph partition \cite{pan2018iterative}, phase correlation \cite{HUANG2021310}, and topological graph \cite{li2020robust} are also introduced to find correspondence. 
Among these methods, using the graphical model with nodes and edges and finding correspondence by matching complete or sub-graphs is a robust and efficient solution since the graphical model can naturally represent 3D topology. The graph matching (GM) method is a robust global strategy. To this end, we present a robust coarse-to-fine registration strategy, with the following main contributions:
\begin{itemize}
    \item A coarse-to-fine registration is achieved by utilizing robust rotation-invariant features and a weighted GM method for estimating the optimal correspondence and transformation parameters in an iterative process.
    \item A new weighted graph matching method is proposed, in which the matching of correspondence is formulated by considering the similarity of nodes and edges in both Euclidean and feature space. The geometric nature is fully considered by building a linear mapping between geometric transformation parameters and correspondence.
\end{itemize}
In the remainder, Sec. II introduces the method and Sec. III gives results and discussions. Then, Sec. IV draws conclusions.

\section{Methodology}
The general workflow of the proposed method is shown in Fig.~\ref{fig:FigX_workflow}. 
First, the keypoints are detected using ISS detector \cite{zhong2009intrinsic} and the rotation-invariant features are generated using RIDF descriptor \cite{huang2020ridf} to encode local contextual information. 
Then, an objective function is formulated to find the optimal correspondence, which maximizes the total similarity between the corresponding graphs formed by keypoints. 
The optimization problem is solved using a fast approximation of the Frank-Wolfe method \cite{wang2019functional} and transformation parameters are estimated in closed form with given correspondence. 
Finally, the correspondence and the transformation parameters are updated and getting closer to the final optimal solution in the iterative process. 
The iterative process will stop when the registration threshold is met.

\subsection{Rotation-invariant feature extraction}
In this section, our previously developed RIDF descriptor encodes the local context around the keypoint. To complete the overall description of our registration method, we start with a brief introduction of the RIDF descriptor.

The local rotation-invariant features can be extracted using the RIDF descriptor. The procedures of the descriptor can be summarized as three steps: (i) calculate normal vector at each keypoint position and voxelize the local surroundings of each keypoint; (ii) compute spherical harmonic (SH) HOG field; (iii) compute regional features and generate the final rotation-invariant features. The details are presented as follows.

For each voxel, the normal vector is calculated to represent the 3D gradients of each voxel and denoted as $\mathbf{d}$. The corresponding spherical coordinate representation of the gradient can be represented as $(\|\mathbf{d}\|,\theta_d,\phi_d)$. Then, the spherical harmonics (SHs) can be calculated as:
\begin{equation}
    \textbf{F}^l_m = \frac{2l+1}{4\pi}\|\mathbf{d}\|\mathbf{Y}^l_m(\theta_d, \phi_d),
\end{equation}
where $\mathbf{Y}_l^m$ is the Schmidit semi-normailzied SHs, which is a function of the degree, order, and the spherical coordinates. Since the SH representation only considers limited degree and order, it can be regraded as a low-frequency filtering.

The SH HOG field can be obtained by normalized the SH coefficients:
\begin{equation}
    \tilde{\mathbf{F}}^l_m = \mathbf{F}_m^l/\sqrt{\Vert\mathbf{d}^2\Vert * K},
\end{equation}
where $K$ is the kernel for the local normalization.




Since the SH HOG representation is a rank-$l$ spherical tensor field, the rotations can be calculated using the Wigner-D matrix $\mathcal{D}^l$. Thus, it can be derived as:
\begin{equation}
    g\tilde{\mathbf{F}}^l = [\Vert \mathbf{d} \Vert\mathcal{D}_g^l\mathbf{Y}^l(\mathbf{d})]\circ \mathbf{T}^g = \mathcal{D}_g^l[\tilde{\mathbf{F}}^l\circ\mathbf{T}_g]
\end{equation}
Thus, in order to construct rotation-invariant features, spherical tensor operations are still needed.



Based on the attributes of SHs, the rotation-invariant features can be obtained by coupling the filtering output of the same rank as the inner product. Finally, the descriptor can be calculated by
\begin{equation}\label{equ:sgd}
    \mathbf{f}_i = \int tensor{\_}product(\tilde{\mathbf{F}}^l(\mathbf{x}-\mathbf{x}'),\mathbf{G}_n(\mathbf{x}'),k)d\mathbf{x}'
\end{equation}
where $\mathbf{G}_n$ denotes the spherical Gaussian derivatives.


\begin{figure*}[!t]
    \centering
    \includegraphics[width=0.85\textwidth]{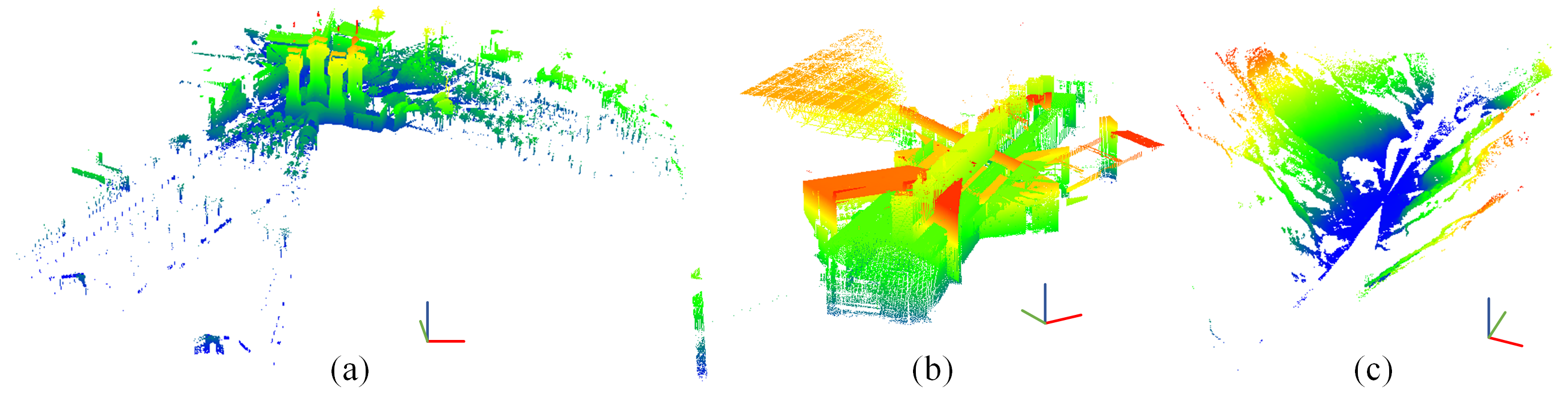}
    \setlength{\abovecaptionskip}{-.25cm}
    \caption{The selected registration pairs as test data. (a) and (b) are from the Resso dataset. (c) is from the WHU-TLS dataset.}
    \label{fig:FigX_dataset}
\end{figure*}
\vspace{-0.5cm}
\begin{table*}[]
    \centering
    \caption{Registration results and comparison with baseline methods}
    \begin{tabular}{ccccccc}
    \toprule
         \multirow{2}{*}{Methods} & \multicolumn{2}{c}{Resso-Outdoor} & \multicolumn{2}{c}{Resso-Indoor} & \multicolumn{2}{c}{WHU-TLS (Mountain)} \\
         & Rotation error & Translation error & Rotation error & Translation error & Rotation error & Translation error\\
         \midrule
         PLADE \cite{chen2019plade} & 0.0810 & 0.0854 & 0.4891 & 0.0168 & \multicolumn{2}{c}{/} \\
         HMMR \cite{dong2020registration}  & \multicolumn{2}{c}{/} & \multicolumn{2}{c}{/} &0.0495 & 0.0180 \\
         GRPC \cite{HUANG2021310} & 0.0727 & 0.3075  & 0.6563 & 0.1833 & 0.2338 & 0.4010\\
         RIDF \cite{huang2020ridf} & 0.4130 & 0.4619  & 0.7482 & 0.2057 & 0.3034 & 0.1573\\
         Our & 0.0812 & 0.0041 & 0.1789 & 0.0008 & 0.0024 & 0.0781\\
         \bottomrule
    \end{tabular}
    \label{tab:regi_results}
\end{table*}

\subsection{GM with geometric transformation}
In this section, a general overview of GM is presented and followed by an optimized solution specialized for matching keypoints, in which the geometric nature of point clouds is involved.
\subsubsection{An overview of GM problem}
GM-based methods achieve pursuing the optimal correspondence between graphs by constructing functions that considers the similarity between nodes and edges of two graphs. Here, we briefly introduce the concept of GM. Given two undirected graphs $\mathcal{G}_1=\left\{V_1, E_1\right\}$ and $\mathcal{G}_2=\left\{V_2, E_2\right\}$ of size $m$ and $n$ respectively, which is defined by a set of nodes $V$ and a set of edges $E$, the task is to find an optimal correspondence $\mathbf{C}\in \left\{0,1\right\}^{m\times n}$ that yields the maximal similarity between the two graphs considering both the node and edge similarity. It can be expressed as the following equation:
\begin{equation}
\begin{aligned}
    \mathbf{C}^* & = \arg\max_\mathbf{C}{\mathbf{C}^T\mathbf{M}\mathbf{C}},
\end{aligned}
\end{equation}
where $\mathbf{M}\in \mathcal{R}^{mn\times mn}$ is the affinity matrix of the two graphs, which measures how well each pair in $\mathcal{G}_1$ matches each pair in $\mathcal{G}_2$. However, the affinity matrix $\mathbf{M}$ is extremely large brings considerable space complexity when the number of nodes in two graphs is large.

The GM problem can also be formulated as an objective function which minimizes the summation of node dissimilarity (unary term) and edge dissimilarity (pairwise term) instead of using $\mathbf{M}$:
\begin{equation}
    \mathbf{C}^* = \arg\max_\mathbf{C}\left\{ \left\langle \mathbf{C},\mathbf{D}\right\rangle_F+\alpha_1\|\mathbf{A}_1-\mathbf{C}\mathbf{A}_2\mathbf{C}^T\|^2_F\right\}
\end{equation}
where $\mathbf{D} \in \mathcal{R}^{m \times n}$ measures the dissimilarity between nodes of the two graphs, and $\mathbf{A}_1 \in \mathcal{R}^{m \times m}$ and $\mathbf{A}_2 \in \mathcal{R}^{n \times n}$ are the adjacency weight matrices of the two graphs. $\alpha_1$ balances the unary term and the pairwise term. $\left\langle\cdot,\cdot\right\rangle$ is the Forbenius dot-product and $\| \cdot \|^2_{F}$ is the Frobenius norm.

This objective function can be solved using constrained optimization methods and the result can be binarized using a post-discretization, such as Hungarian algorithm \cite{kuhn1955hungarian}.


\subsubsection{Problem formulation}
Although the original GM methods can provide good solutions for graphs embedded in explicit or implicit feature spaces, the geometric nature of the real data is not fully considered when we apply this type of method to point cloud registration. Inspired by FRGM \cite{wang2019functional}, we improve the graphical model by building connections between correspondence with geometric transformation parameters. 

Assuming that the geometric transformation $T$ can be represented by a linear representation map of $\mathbf{C}$:
\begin{equation}
    T(V_1) = \mathbf{C}V_2.
\end{equation}

Since the geometric structure does not change during the geometric transformation, the edge length between the original edge and the corresponding edge is preserved. In addition, the edge length can be presented in two space domains, namely the feature domain and the spatial domain. The original edge length of feature space and Euclidean space can be written as $V^{fd}_{1,i1i2}$ and $V^{ed}_{1,i1i2}$, where $1$ represent the graph number and $i1i2$ represents edge formed by nodes $i1$ and $i2$. The transformed edge length $T(V^{fd}_{1,i1i2})$ and $T(V^{ed}_{1,i1i2})$ can be represented using $\mathbf{C}$ as $(\mathbf{C}V^{fd}_2)_{i1i2}$ and $(\mathbf{C}V^{ed}_2)_{i1i2}$. Thus, the objective function can be reformulated by substituting the pairwise term:
\begin{equation}
\begin{aligned}
    J_1(\mathbf{C}) &= \left\langle\mathbf{C},\mathbf{D}\right\rangle_F \\
    &+ \alpha_1\sum_{(i1,i2)}\mathbf{A}^{fd}_{1,i1i2}(\|\mathbf{V}^{fd}_{1,i1i2}\|-\|(\mathbf{C}V^{fd}_2)_{i1i2}\|)^2 \\
    &+ \alpha_2\sum_{(i1,i2)}\mathbf{A}^{ed}_{1,i1i2}(\|\mathbf{V}^{ed}_{1,i1i2}\|-\|(\mathbf{C}V^{ed}_2)_{i1i2}\|)^2
\end{aligned}
\end{equation}





However, since $J$ is nonconvex, the solution for the objective function always reaches a local minimum. If a post-discretization is directly conducted on the initial result, the final result will be of low quality. 

\subsubsection{Model optimization}
Assuming that there is an offset between $T(V_{1,i})$ and its correct match $V_{2,\sigma_i}$ and the initial match is $\mathbf{C}_1^*$. To consider the offset, the differences between adjacency offset vectors are minimized. Thus, the correspondence is further optimized using the objective function shown as follows:
\begin{equation} \label{equ:optimization}
\begin{aligned}
     J_2(\mathbf{C}) = & \left\langle\mathbf{C},\mathbf{D}\right\rangle_F \\
    & +  \alpha_3\tr((\mathbf{C}V_2-\mathbf{C}^*_1V_2)^T\mathbf{L}_{\mathbf{A}_t}(\mathbf{C}V_2-\mathbf{C}_1^*V_2))
\end{aligned}
\end{equation}
where $\mathbf{L}_{\mathbf{A}_t} = \diag(\mathbf{A}_t\mathbf{I}-\mathbf{A}_t)$ and $\mathbf{A}_t$ indicates the adjacency matrix between the transformed nodes. $\alpha_3$ is a weight factor.

Since we aim to solve 7DOF transformation, the transformation can be written as:
\begin{equation}
    T(V) = sV\mathbf{R} + \mathbf{t}
\end{equation}
where $s$, $\mathbf{R}$, and $\mathbf{t}$ denote the scaling factor, the rotation matrix, and the translation.

With the estimated correspondence $\mathbf{C}$, the parameters of the transformation can be calculated by solving the following objective function:
\begin{equation}
\begin{aligned}
    J_3(T) = &\sum_i\|V_{1,i}-T^{-1}((\mathbf{C}V_2)_i)\|^2 \\
    &+\alpha_3\sum_{(i1,i2)}\mathbf{A}_{1,i1i2}\|(V_{1,i1i2})-T^{-1}((\mathbf{C}V_2)_{i1i2})\|^2
\end{aligned}
\end{equation}
\subsubsection{Iterative process}
Although the correspondence and the transformation parameters can be estimated by solving the aforementioned objective function, the result is still not robust enough. Thus, a fast approximated FW method is utilized to gradually updating the correspondence and transformation parameters iteratively. Once the transformation parameters are obtained, the correspondence can be optimized using Eq.~\ref{equ:optimization}. The iterative process will finally stop when the change of registration errors is less than the tolerance or the iteration number reaches the defined maximum value.
\section{Experiments and results}

\subsection{Experimental data}
The proposed registration method was tested using three TLS point cloud pairs from two benchmark datasets. Specifically, two registration pairs were selected from the Resso dataset, including one registration pair from the outdoor scene and one from the indoor scene. Another registration pair is from the WHU-TLS dataset \cite{dong2020registration}, obtained from a mountain area (see Fig.~\ref{fig:FigX_dataset}). The performance of the proposed registration method was evaluated in terms of rotation errors and translation errors.
\subsection{Experimental results}
\begin{figure*}[t]
    \centering
    \includegraphics[width=0.8\textwidth]{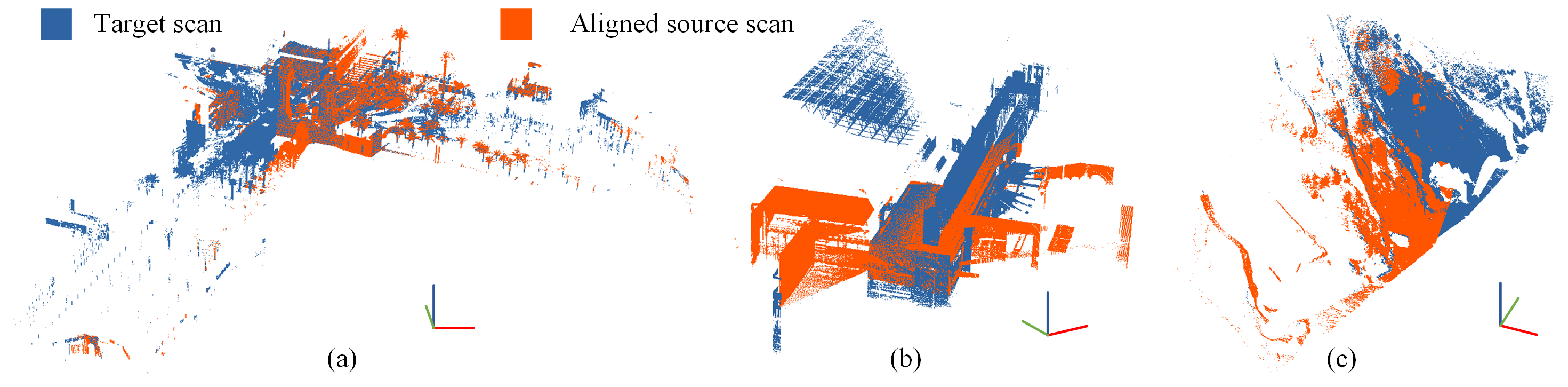}
    \caption{Illustration of the registration results using the proposed registration method. (a) and (b) are registered point clouds of the Resso dataset. (c) Registered point clouds of the WHU-TLS dataset.}
    \label{fig:FigX_Results}
\end{figure*}
\subsubsection{Comparison with other methods}
The registration results of three registration pairs are listed in Table~\ref{tab:regi_results}. To verify the performance of our method, the registration results are compared with several baseline methods. PLADE \cite{chen1992object} and Hierarchical merging based multiview registration (HMMR) \cite{dong2020registration} are the baseline methods proposed by the publishers of the Resso and the WHU-TLS datasets, respectively. PLADE used the strong constraint provided by planes and designed a new plane-based descriptor. HMMR provided a hybrid solution that combined global and local features for multiscan registration. GPRC registered point clouds by finding global correspondence in the frequency domain using a phase correlation-based method \cite{HUANG2021310}. RIDF used the RIDF features and estimated transformation parameters using a RANSAC estimator \cite{huang2020ridf}. 
As shown in Table \ref{tab:regi_results}, for the Resso-Outdoor dataset, our proposed method can provide results with the rotation error of 0.08 degree and the translation error of 4 mm. Our method can reach the same level results as the baseline method (PLADE) and outperforms other methods, such as GRPC and RIDF. As for the Resso-Indoor dataset, our proposed method can obtain much better registration results than the other methods, achieving less than a 0.2-degree rotation error and less than 1-mm translation error. The last dataset covers a non-urban area, which raises great challenges for many registration methods based on strong geometric constraints. It can be seen that our method also produces satisfying results with 0.002-degree rotation error and 78-mm translation error. Although the translation error is larger than that of results obtained by HMMR, the rotation error is of a smaller value. Compared with HMMR, our proposed method can reach almost the same level of registration results. Besides, our method performed better in the task of registration compared with GRPC and RIDF. In general, in terms of both the rotation and translation accuracy, our method outperforms other methods. The accuracy of registration results almost reaches the level of fine registration.
\subsubsection{Analysis of sensitivity to noise}
\begin{figure*}[t]
    \centering
    \includegraphics[width=1.0\textwidth]{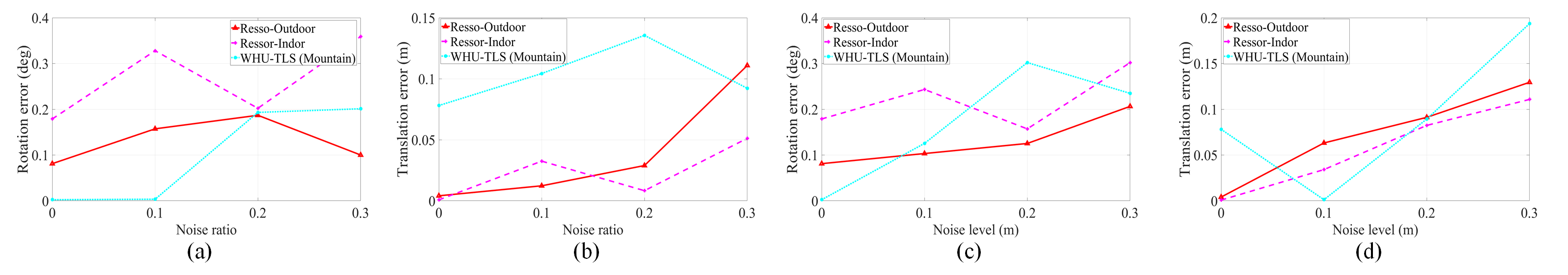}
    \caption{Analysis of sensitivity to noise. (a) Rotation errors under different noise ratios. (b) Translation error under different noise ratios. (c) Rotation errors under different noise levels. (d) Translation error under different noise levels.}
    \label{fig:FigX_sensitivity}
\end{figure*}
The proposed method's sensitivity to noise is tested by adding Gaussian noise to the original point clouds and varying the noise ratio and noise level. The noise ratio rose from 0.1 to 0.3. As for the noise level, it was set from 0.1 m to 0.3 m. The registration results of the three registration pairs are shown in Fig.~\ref{fig:FigX_sensitivity}. As seen from Fig.~\ref{fig:FigX_sensitivity}a and Fig.~\ref{fig:FigX_sensitivity}b, the performance of our method is influenced by the change of noise ratio. Considering both rotation and translation errors, all three datasets' registration results are getting worse with the increment of noise ratio but vary at an acceptable level. Additionally, it can be seen that the matching of the WHU-TLS dataset is comparatively more sensitive to the change of noise ratios. The change of registration results under different noise levels are shown in Fig.~\ref{fig:FigX_sensitivity}c and Fig.~\ref{fig:FigX_sensitivity}d. It can be seen that the increase of noise level actually influences the registration results with higher rotation errors and translation errors. The WHU-TLS dataset also shows higher sensitivity to the change of noise levels. From the figures, it can be concluded that our method is kind of sensitive to the change of noise, but the registration results remain of high accuracy. Compared with datasets covering regular-shaped areas, such as urban or indoor areas, registration of non-regular-shaped areas using the proposed method shows higher sensitivity.

\vspace{-0.25cm}
\section{Conclusion}
In this letter, we propose a coarse-to-fine registration method, which uses rotation-invariant features for local context encoding using RIDF descriptor and a weighted GM-based method for global optimization. The RIDF descriptor achieves robust rotation-invariant feature extraction using Fourier-based operations. The GM method iteratively finds correspondence between keypoints considering similarity between nodes and edges in both feature and Euclidean spaces. The geometric nature is considered by representing transformation using a linear representation map of correspondence. The proposed method was tested under various scenarios. The experimental results indicated that our method had good performance in the registration task and was robust to noise. In future work, the deep-learning-based method can be utilized to solve the GM problem.

\ifCLASSOPTIONcaptionsoff
  \newpage
\fi



%
\bibliographystyle{IEEEtran}
\bibliography{GM_Ref}

%

\end{document}